\documentclass{article} 
\usepackage{nips12submit_e,times}
\usepackage{amsmath, amsthm, amssymb}
\usepackage[mathcal]{eucal}
\usepackage{graphicx}
\usepackage{verbatim}
\usepackage[sort]{natbib}
\usepackage{booktabs}

\title{Discriminative Recurrent Sparse Auto-Encoders}

\author{
Jason Tyler Rolfe \& Yann LeCun \\ 
Courant Institute of Mathematical Sciences, New York University\\ 
719 Broadway, 12th Floor \\
New York, NY 10003\\
\texttt{\{rolfe, yann\}@cs.nyu.edu} \\
}

\nipsfinalcopy 

\newcommand{\bv}{\mathbf{b}}

\newcommand{\hid}{\mathbf{z}}
\newcommand{\nobfhid}{z}
\newcommand{\inp}{\mathbf{x}}
\newcommand{\nobfinp}{x}
\newcommand{\out}{y}
\newcommand{\C}{\mathbf{C}}
\newcommand{\D}{\mathbf{D}}
\newcommand{\E}{\mathbf{E}}
\newcommand{\I}{\mathbf{I}}

\newcommand{\Sm}{\mathbf{S}}
\newcommand{\logistic}{\text{logistic}}

\begin{document}


\maketitle

\begin{abstract}
We present the discriminative recurrent sparse auto-encoder model, comprising a recurrent encoder of rectified linear units, unrolled 
for a fixed number of iterations, and connected to two linear decoders that reconstruct the input and predict its supervised classification. 
Training via backpropagation-through-time initially minimizes an unsupervised sparse reconstruction error; the loss function is then augmented with a discriminative term on the supervised classification.  
The depth implicit in the temporally-unrolled form allows the system to exhibit far more representational power, while keeping the number of trainable parameters fixed.

From an initially unstructured network the hidden units differentiate into categorical-units, each of which represents an input prototype with a well-defined class; and part-units representing deformations of these prototypes.  The learned organization of the recurrent encoder is hierarchical: part-units are driven directly by the input, whereas the activity of categorical-units builds up over time through interactions with the part-units.  Even using a small number of hidden units per layer, discriminative recurrent sparse auto-encoders achieve excellent performance on MNIST. 
\end{abstract}

\section{Introduction}

Deep networks complement the hierarchical structure in natural data~\citep{bengio2009}.  
By breaking complex calculations into many steps, deep networks can gradually build up complicated decision boundaries or input transformations, facilitate the reuse of common substructure, and explicitly compare alternative interpretations of ambiguous input~\citep{lee2008, zeiler2011}.  
Leveraging these strengths, deep networks have facilitated significant advances in solving sensory problems like visual classification and speech recognition~\citep{hinton2006, dahl2012, hinton2012}.  

Although deep networks have traditionally used independent parameters for each layer, they are equivalent to recurrent networks in which a disjoint set of units is active on each time step.  
The corresponding representations are sparse, and thus invite the incorporation of powerful techniques from sparse coding~\citep{olshausen1996, olshausen1997, ranzato2006, lee2008, glorot2011}.
Recurrence opens the possibility of sharing parameters between successive layers of a deep network. 

This paper introduces the \emph{Discriminative Recurrent Sparse Auto-Encoder} model (DrSAE), comprising a recurrent encoder of rectified linear units~\citep[ReLU; ][]{salinas1996, jarrett2009, nair2010, glorot2011, coates2011}, connected to two linear decoders that reconstruct the input and predict its supervised classification.  The recurrent encoder is unrolled in time for a fixed number of iterations, with the input projecting to each resulting layer, and trained using backpropagation-through-time~\citep{rumelhart1986}.  Training initially minimizes an unsupervised sparse reconstruction error; the loss function is then augmented with a discriminative term on the supervised classification.  
In its temporally-unrolled form, the network can be seen as a deep network, with parameters shared between the hidden layers. The temporal depth allows the system to exhibit far more representational power, while keeping the number of trainable parameters fixed.

Interestingly, experiments show that DrSAE does not just discover more discriminative ``parts'' of the form conventionally produced by sparse coding. Rather, the hidden units spontaneously differentiate into two types: a small number of categorical-units and a larger number of part-units. The categorical-units have decoder bases that look like prototypes of the input classes. They are weakly influenced by the input and activate late in the dynamics as the result of interaction with the part-units. In contrast, the part-units are strongly influenced by the input, and encode small transformations through which the prototypes of categorical-units can be reshaped into the current input. 
Categorical-units compete with each other through mutual inhibition and cooperate with relevant part-units. This can be interpreted as a representation of the data manifold in which the categorical-units are points on the manifold, and the part-units are akin to tangent vectors along the manifold.  

\subsection{Prior work}

The encoder architecture of DrSAE is modeled after the Iterative Shrinkage and Threshold Algorithm (ISTA), a proximal method for sparse coding~\citep{chambolle1998, daubechies2004}.  \citet{gregor2010} showed that the sparse representations computed by ISTA can be efficiently approximated by a structurally similar encoder with a less restrictive, learned parameterization.  Rather than learn to approximate a precomputed optimal sparse code, the LISTA autoencoders of \citet{sprechmann2012a, sprechmann2012b} are trained to directly minimize the sparse reconstruction loss function.  DrSAE extends LISTA autoencoders with a non-negativity constraint, which converts the shrink nonlinearity of LISTA into a rectified linear operator; and introduces a unified classification loss, as previously used in conjunction with traditional sparse coders~\citep{bradley2008, mairal2009, mairal2012} and other autoencoders~\citep{ranzato2008, boureau2010}.

DrSAEs resemble the structure of deep sparse rectifier neural networks \citep{glorot2011}, but differ in that the parameter matrices at each layer are tied \citep{bengio2012a}, the input projects to all layers, and the outputs are normalized.  
DrSAEs are also reminiscent of the recurrent neural networks investigated by \citet{bengio1996}, but use a different nonlinearity and a heavily regularized loss function.  Finally, they are similar to the recurrent networks described by \citet{seung1998}, but have recurrent connections amongst the hidden units, rather than between the hidden units and the input units, and introduce classification and sparsification losses.

\section{Network architecture}

\begin{figure}[tb]
  \begin{center}
    \includegraphics[width=5.5in, keepaspectratio=true]{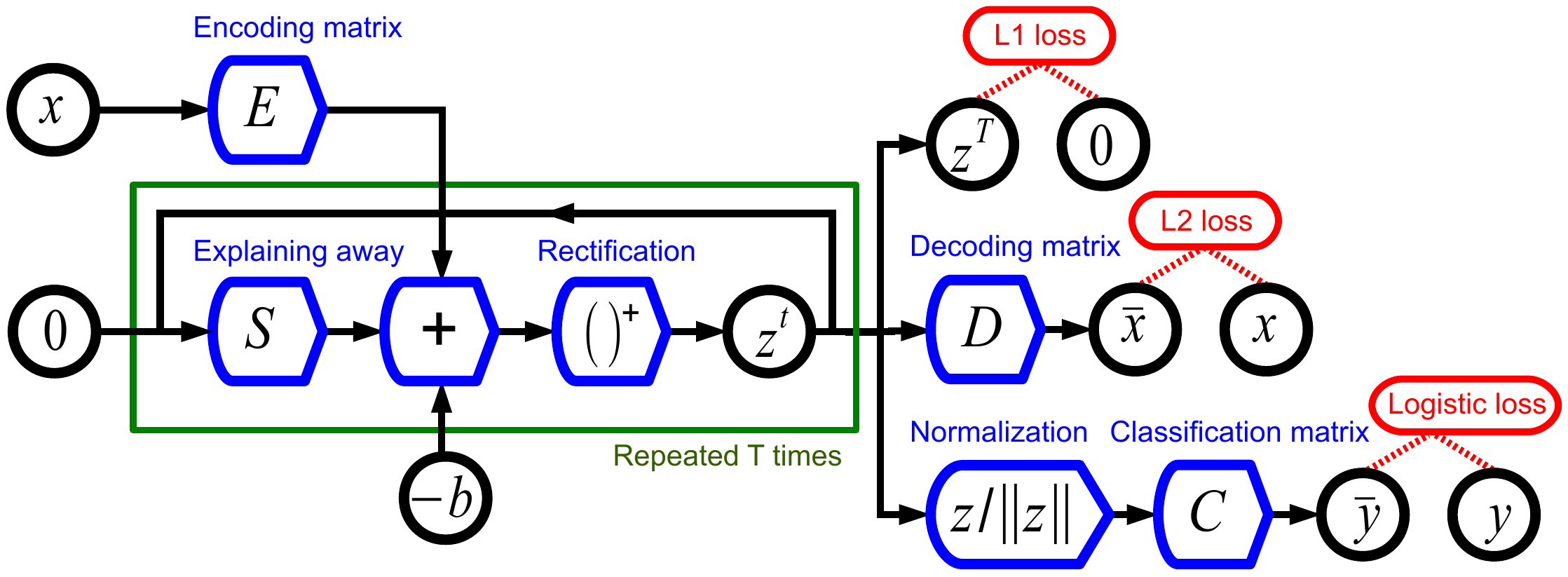}
  \end{center}
  \caption{The discriminative recurrent sparse auto-encoder (DrSAE) architecture.  $\hid^t$ is the hidden representation after iteration $t$ of $T$, and is initialized to $\hid^0 = 0$; $\inp$ is the input; and $\mathbf{\out}$ is the supervised classification.  Overbars denote approximations produced by the network, rather than the true input.  $\E$, $\Sm$, $\D$, and $\bv$ are learned parameters.  \label{architecture_figure}} 
\end{figure}


In the following, we use lower-case bold letters to denote vectors, upper-case bold letters to denote matrices, superscripts to indicate iterative copies of a vector, and subscripts to index the columns (or rows, if explicitly specified by the context) of a matrix or (without boldface) the elements of a vector.
We consider discriminative recurrent sparse auto-encoders (DrSAEs) of rectified linear units with the architecture shown in figure~\ref{architecture_figure}: 
\begin{equation} \label{layer-dynamics}
\hid^{t+1} = \max\left(0, \E \cdot \inp + \Sm \cdot \hid^t - \bv \right)
\end{equation}
for $t = 1, \ldots, T$, where $n$-dimensional vector $\hid^t$ is the activity of the hidden units at iteration $t$, 
$m$-dimensional vector $\inp$ is the input, and $\hid^{t=0} = 0$.  Unlike traditional recurrent autoencoders \citep{bengio2012a}, the input projects to every iteration.  We call the $n \times m$ parameter matrix $\E$ the encoding matrix, and the $n \times n$ parameter matrix $\Sm$ the explaining-away matrix.
The $n$-element parameter vector $\bv$ contains a bias term.  The parameters also include the $m \times n$ decoding matrix $\D$ and the $l \times n$ classification matrix $\C$.

We pretrain DrSAEs using stochastic gradient descent on the unsupervised loss function
\begin{equation} \label{reconstruction-loss}
L^U = \frac{1}{2} \cdot \left| \left| \inp - \D \cdot \hid^T \right| \right|_2^2 + \lambda \cdot \left|\left|\hid^T \right|\right|_1,
\end{equation}  
with the magnitude of the columns of $\D$ bounded by $1$,\footnote{This sets the scale of $\hid$; otherwise, the magnitude of $\hid$ will shrink to zero and the magnitude of the columns of $\D$ will explode.  This and all other such constraints are enforced by a projection after each SGD step.} and the magnitude of the rows of $\E$ bounded by $\frac{1.25}{T}$.\footnote{The size of each ISTA step must be sufficiently small to guarantee convergence.  As the step size grows large, the input will be over-explained by multiple aligned hidden units, leading to extreme oscillations.  This bound serves the same function as $\ell_2$ weight regularization \citep{hinton2010}.  The particular value of the bound is heuristic, and was determined by an informal search of parameter space.}  We then add in the supervised classification loss function
\begin{equation} \label{discriminative-loss}
L^S = \logistic_\out \left( \C \cdot \frac{\hid^T}{\left|\left| \hid^T \right|\right|} \right) \, ,
\end{equation}
where the multinomial logistic loss function is defined by 
\begin{equation*}
  \logistic_\out(\hid) = \nobfhid_\out - \log\left( \sum_i e^{\nobfhid_i} \right) \, ,
\end{equation*}
and $\out$ is the index of the desired class.\footnote{Consistent with standard autoencoders but unlike traditional applications of backpropagation-through-time, the loss functions $L^U$ and $L^S$ only depend directly on the final iteration of the hidden units $\hid^T$.}
Starting with the parameters learned by the unsupervised pretraining, we perform discriminative fine-tune by stochastic gradient descent on $L^U + L^S$, with the magnitude of the rows of $\C$ bounded by $5$.\footnote{As in the case of the encoder, this serves the same function as $\ell_2$ weight regularization \citep{hinton2010}.  The particular value of the bound is heuristic, and was determined by an informal search of parameter space.}  
The learning rate of each matrix is scaled down by the number of times it is repeated in the network, and the learning rate of the classification matrix is scaled down by a factor of $5$, to keep the effective learning rate consistent amongst the parameter matrices.

\begin{figure}[tbp]
  \begin{center}
    \begin{tabular}{p{0.10in}p{5.2in}}
      \parbox[b]{0in}{(a) \\ Enc \vspace{0.9cm}} & \includegraphics[width=5.2in, keepaspectratio=true]{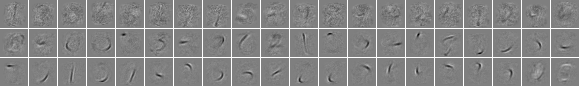} \\
      \parbox[b]{0in}{(b) \\ Dec \vspace{0.9cm}} & \includegraphics[width=5.2in, keepaspectratio=true]{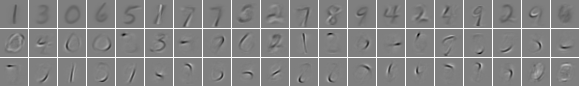} \\
      \parbox[b]{0in}{(c) \\ Clas \vspace{0.2cm}} & \includegraphics[width=5.2in, keepaspectratio=true]{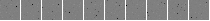} \\
    \end{tabular}
  \end{center}
  \caption{The hidden units differentiate into spatially localized part-units, which have well-aligned encoders and decoders; and global prototype categorical-units, which have poorly aligned encoders and decoders.  A subset of the rows of encoding matrix~$\E$~(a) and the columns of decoding matrix~$\D$~(b), and all rows of the classification matrix~$\C$~(c) after training. The first row of (a,b) shows the most categorical units; the last row contains the least categorical units; and the middle row evenly steps through the remaining units in order of categoricalness.  Gray pixels denote connections with weight~$0$; darker pixels indicate more positive connections.\label{dictionary_figure}} 
\end{figure}

We train DrSAEs with $T=11$ recurrent iterations (ten nontrivial passes through the explaining-away matrix~$\Sm$)\footnote{The chosen number of recurrent iterations achieves a heuristic balance between representational power and computational expense.  Experiments were conducted with $T \in \left\{ 2,6,11,21 \right\}$.} and $400$~hidden units on the MNIST dataset of $28 \times 28$ grayscale handwritten digits \citep{lecun1998}, with each input normalized to have $\ell_2$ magnitude equal to $1$.  We use a training set of 50,000 elements, and a validation set of 10,000 elements to perform early-stopping.
Encoding, decoding, and classification matrices learned via this procedure are depicted in figure~\ref{dictionary_figure}.

The dynamics of equation~\ref{layer-dynamics} are inspired by the Learned Iterative Shrinkage and Thresholding Algorithm (LISTA)~\citep{gregor2010}, an efficient approximation to the sparse coding Iterative Shrinkage and Threshold Algorithm (ISTA)~\citep{chambolle1998, daubechies2004}.
ISTA is an 
algorithm for minimizing the $\ell_1$-regularized reconstruction loss function $L^U$ of equation~\ref{reconstruction-loss} 
with respect to $\hid^T$.  
It is defined by the iterative step
\begin{equation*}
\hid^{t+1} = h_{\alpha \cdot \lambda} \left(\alpha \cdot \D^{\top} \cdot \inp + \left(\I - \alpha \cdot \D^{\top} \cdot \D\right) \cdot \hid^t \right) \, ,
\end{equation*}  
where $\left[h_{\theta}(\mathbf{x}) \right]_i = \text{sign}\left(x_i \right) \cdot \max\left(0, \left|x_i \right| - \theta \right)$ and $\alpha$ is a small step-size parameter.
With non-negative units, ISTA is equivalent to projected gradient descent of $L^U$ of equation~\ref{reconstruction-loss}.
As the number of iterations $T \rightarrow \infty$, a DrSAE defined by equation~\ref{layer-dynamics} becomes a non-negative version of ISTA if it satisfies the restrictions:
\begin{equation} \label{ISTA-restriction}
\E = \alpha \cdot \D^{\top} , \hspace{1cm}
\Sm = \I - \alpha \cdot \D^{\top} \cdot \D \, , \hspace{0.5cm}
b_i= \alpha \cdot \lambda \, , \hspace{0.5cm} 
\text{and} \hspace{0.5cm} \nobfhid_i^t \geq 0 \, ,
\end{equation}
where the positive scale factor $\alpha$ is less than the maximal eigenvalue of $\D^{\top} \cdot \D$, and $\I$ is the $n \times n$ identity matrix.

As in LISTA, but unlike ISTA, the encoding matrix $\E$ and explaining-away matrix $\Sm$ in a DrSAE are independent of the decoding matrix $\D$.
Connections from the input to the hidden units, and recurrent connections between the hidden units, are all-to-all, so the network structure is agnostic to permutations of the input.    
DrSAEs can also be understood as deep, feedforward networks with the parameter matrices tied between the layers.

\section{Analysis of the hidden unit representation} 

\begin{figure}[tb] 
  \begin{center}
    \begin{tabular}{p{0.04in}p{2.4in}p{0.04in}p{2.4in}}
      & \hspace{0.5cm} 11 enc iters, no discriminative training & & \hspace{0.7cm} 2 enc iters, no discriminative training \\
      \parbox[b]{0in}{(a) \vspace{3.8cm}} & \includegraphics[width=2.5in, keepaspectratio=true]{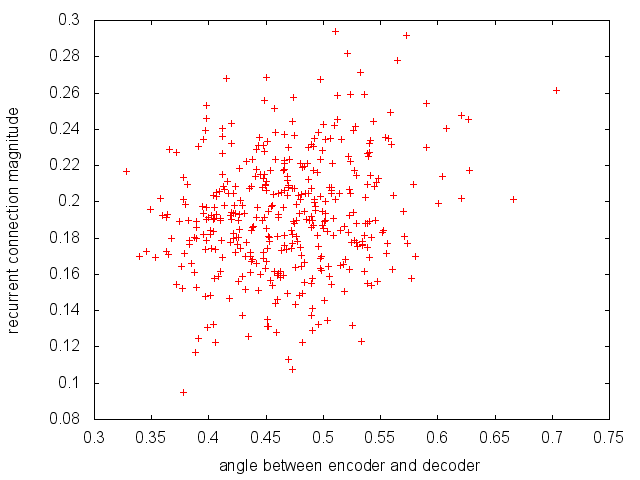} &
      \parbox[b]{0in}{(b) \vspace{3.8cm}} & \includegraphics[width=2.5in, keepaspectratio=true]{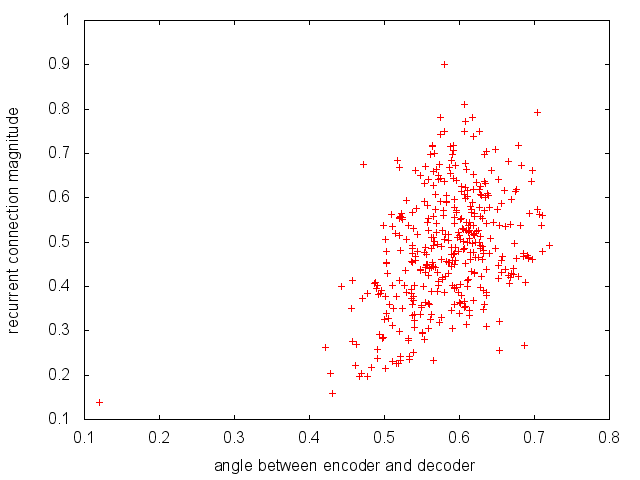} \\
      & \hspace{0.3cm} 11 enc iters, with discriminative training & & \hspace{0.4cm} 2 enc iters, with discriminative training \\
      \parbox[b]{0in}{(c) \vspace{3.8cm}} & \includegraphics[width=2.5in, keepaspectratio=true]{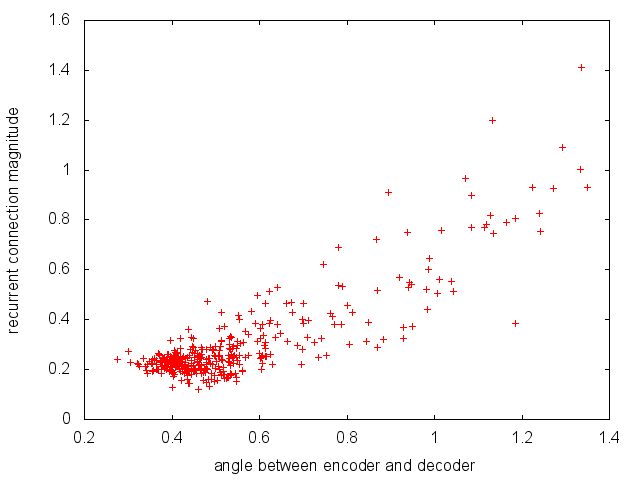} &
      \parbox[b]{0in}{(d) \vspace{3.8cm}} & \includegraphics[width=2.5in, keepaspectratio=true]{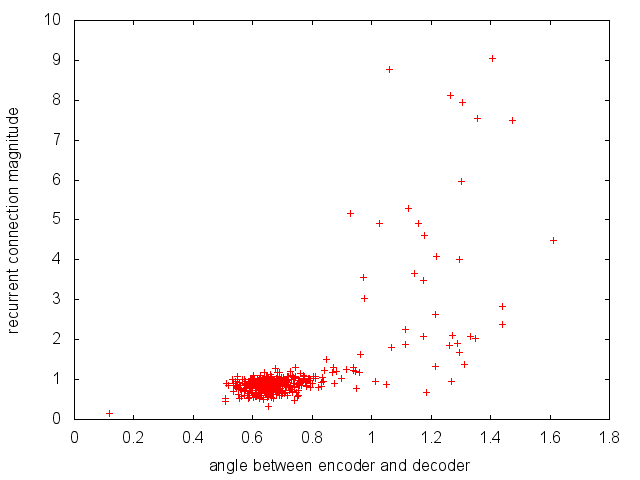} \\
      \parbox[b]{0in}{(e) \vspace{3.8cm}} & \includegraphics[width=2.5in, keepaspectratio=true]{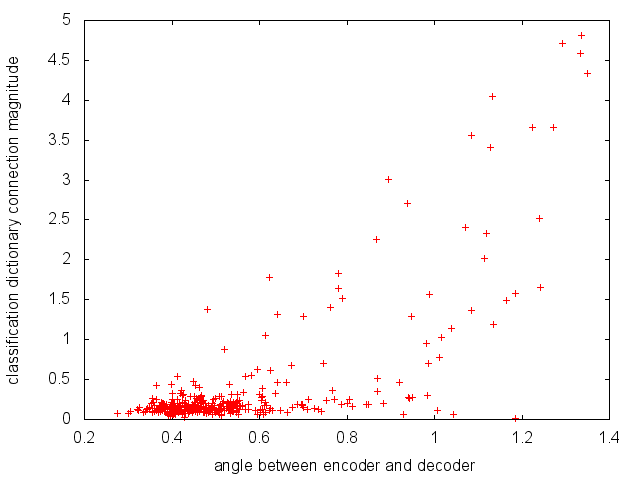} &
      \parbox[b]{0in}{(f) \vspace{3.8cm}} & \includegraphics[width=2.5in, keepaspectratio=true]{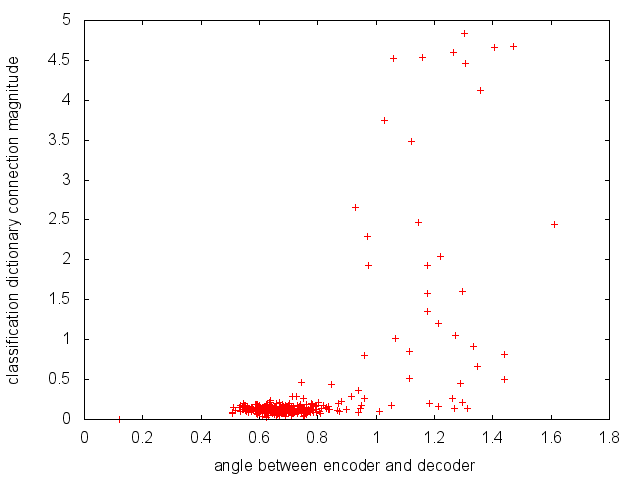}
    \end{tabular}
  \end{center}
  \caption{The hidden units differentiate into two populations after discriminative fine-tuning.  The magnitude of row $\left( \Sm - \I \right)_i$~(a,b,e) and $\C_i$~(c,d), versus the angle between encoder row and decoder column, for each unit from networks using 11 (a,c,e) and 2 (b,d,f) iterations.  All plots are from discriminatively fine-tuned networks except (a,b), which are only subject to unsupervised pretraining.  We call the dense cloud in the bottom-left part-units, and the tail extending to the top-right categorical-units. \label{two_classes_figure}}
\end{figure}

Discriminative fine-tuning naturally induces the hidden units of a DrSAE to differentiate into a hierarchy-like continuum.  On one extreme are part-units, which perform an ISTA-like sparse coding computation;
on the other are categorical-units, which use a sophisticated form of pooling to integrate 
over matching part-units, 
and implement winner-take-all dynamics amongst themselves.  Converging lines of evidence indicate that these two groups use distinct computational mechanisms and serve different representational roles.  

In the ISTA algorithm, each row of the encoding matrix $\E_i$ (which we sometimes call the encoder of unit $i$) is proportional to the corresponding column of the decoding matrix $\D_i$ (which we call the decoder of unit $i$), and each row $\left( \Sm - \I \right)_i$ is proportional to $\left( \D_i \right)^\top \cdot \D$, as in equation~\ref{ISTA-restriction}.  As a result, the angle between $\E_i$ and $\D_i$, and the angle between the rows of $\Sm - \I$ and $\D^{\top} \cdot \D$, are both simple measures of the degree to which a unit's dynamics follow the ISTA algorithm, and thus perform sparse coding.\footnote{We always use $\Sm - \I$ when plotting recurrent connection strength, since it governs the perturbation of the otherwise stable hidden unit activations, as in projected gradient descent of $L^U$; i.e., ISTA.}  These quantities are equal to $0$ in the case of perfect ISTA, and grow larger as the network diverges from ISTA.
Of these two angles, the explaining-away matrix comparison is more difficult to interpret, since a distortion of any one unit's decoding column $\D_i$ will affect all rows of $\D^{\top} \cdot \D$, whereas the angle between the encoder row $\E_i$ and decoder column $\D_i$ only depends upon a single unit. For this reason, we use the angle between the encoder row and decoder column as a measure of the position of each unit on the part/categorical continuum.  

Figure~\ref{two_classes_figure} plots, for each unit $i$, the magnitude of row $\left( \Sm - \I \right)_i$ and column $\C_i$, versus the angle between row $\E_i$ and column $\D_i$.
Before discriminative fine-tuning, there are no categorical-units; the angle between the encoder row and decoder column is small and the incoming recurrent connections are weak for all units, as in figure~\ref{two_classes_figure}(a,b).
After discriminative fine-tuning, there remains a dense cloud of points for which the angle between the encoder row and decoder column is very small, and the incoming recurrent and outgoing classification connections are weak.  Abutting this is an extended tail of points that have a larger angle between the encoder row and decoder column, and stronger incoming recurrent and outgoing classification connections.  We call units composing the dense cloud \emph{part-units}, since they have ISTA-compatible connections, while we refer to those making up the extended tail as \emph{categorical-units}, since they have strong connections to the classification output.\footnote{For the purpose of constructing figures characterizing the difference between part-units and categorical-units, we consider units with encoder-decoder angle less than $0.5$ radians to be part-units, and units with encoder-decoder angle greater than $0.7$ radians to be categorical-units.  These thresholds are heuristic, and fail to reflect the continuum that exists between part- and categorical-units, but they facilitate analysis of the extremes.}  When trained on MNIST, part-units have localized, pen stroke-like decoders, as can be seen in the bottom rows of figure~\ref{dictionary_figure}(a,b).  Categorical-units, in contrast, tend to have whole-digit prototype-like decoders, as in the top rows of figure~\ref{dictionary_figure}(a,b).   Discriminative fine-tuning induces the differentiation of categorical-units regardless of the depth of the encoder.


\subsection{Part-units}

\begin{figure}[tb] 
  \begin{center}
    \begin{tabular}{p{0.04in}p{2.4in}p{0.04in}p{2.4in}}
      \parbox[b]{0in}{(a) \vspace{3.8cm}} & \includegraphics[width=2.5in, keepaspectratio=true]{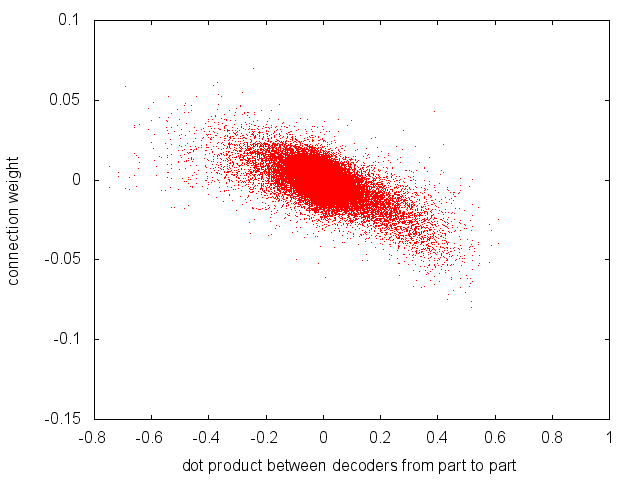} &
      \parbox[b]{0in}{(b) \vspace{3.8cm}} & \includegraphics[width=2.5in, keepaspectratio=true]{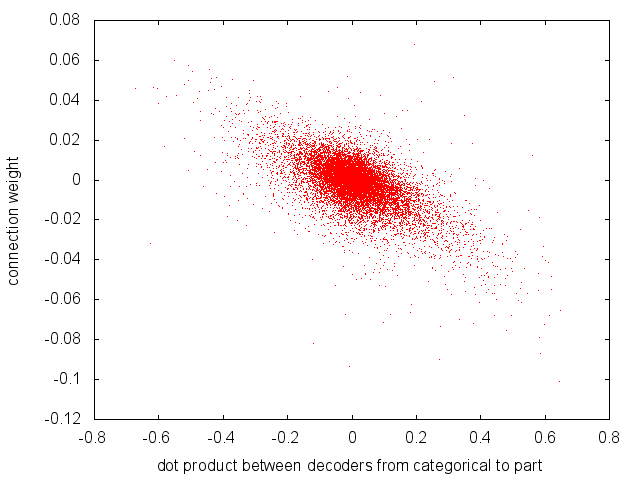}  \\
      \parbox[b]{0in}{(c) \vspace{3.8cm}} & \includegraphics[width=2.5in, keepaspectratio=true]{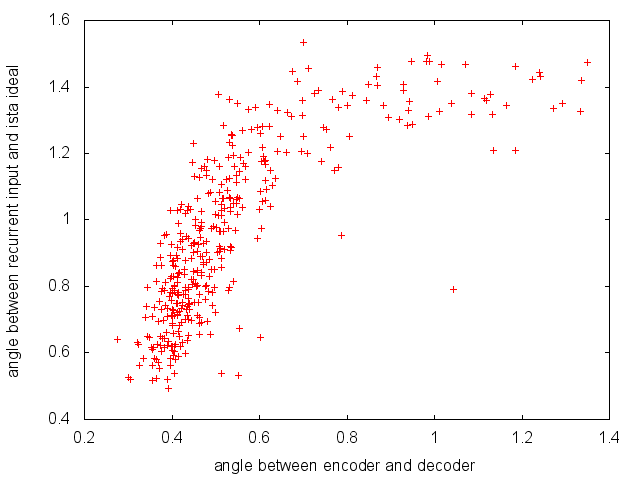}
    \end{tabular}
  \end{center}
  \caption{Part-units have connections consistent with ISTA.  The actual connection weights $\Sm - \I$ versus the ISTA-predicted weights $\D^{\top} \cdot \D$, for connections from part-units to part-units~(a) and categorical-units to part-units~(b); and the angle between the rows of $\Sm - \I$ and the ISTA-ideal $\D^{\top} \cdot \D$ versus the angle between the encoder rows and decoder columns~(c).  Units are considered part-units if the angle between their encoder and decoder is less than $0.5$ radians, and categorical-units if the angle between their encoder and decoder is greater than $0.7$ radians. \label{ista_ideal_connections_figure}}
\end{figure}

Examination of the relationship between the elements of $\Sm - \I$ and $\D^{\top} \cdot \D$ confirms that part-units with an encoder-decoder angle less than $0.5$ radians abide by ISTA, and so perform sparse coding on the residual input after the categorical-unit prototypes are subtracted out.  
The prominent diagonals with matching slopes in figure~\ref{ista_ideal_connections_figure}(a,b), which plot the value of $S_{i,j} - \delta_{i,j}$ versus $\D_i \cdot \D_j$ for connections between part-units, and from categorical-units to part-units, respectively, demonstrate that part-units receive ISTA-consistent connections from all units.  The fidelity of these connections to the ISTA ideal is not strongly dependent upon whether the afferent units are ISTA-compliant part-units, or ISTA-ignoring categorical-units.  As a result, the part-units treat the categorical-units as if they were also participating in the reconstruction of the input, and only attempt to reconstruct the residual input not explained by the categorical-unit prototypes.  

\begin{figure}[bt] 
  \begin{center}
    \begin{tabular}{p{5.5in}}
      \hspace{-0.1in} \begin{tabular}{p{0.175in}p{4in}} Dest & Source units \end{tabular} \\
      \includegraphics[width=5.4in, keepaspectratio=true]{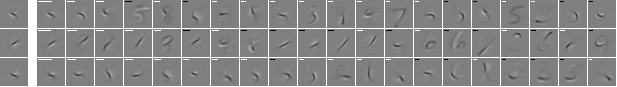} 
    \end{tabular}
  \end{center}
  \caption{Part-units receive ISTA-compatible connections and thus perform sparse coding on the residual input after the contribution of the categorical-units is subtracted out.  The decoders of the twenty units with the strongest explaining-away connections $\left|S_{i,j} - \delta_{i,j}\right|$ to three typical part-units, sorted by connection magnitude.  The left-most column depicts the decoder of the recipient part-unit.  The bars above the decoders in the remaining columns indicate the strength of the connections.  Black bars are used for positive connections, and white bars for negative connections.  \label{incoming_part_figure}}
\end{figure}

As can be seen in figure~\ref{ista_ideal_connections_figure}(c), the degree to which the encoder conforms to the ISTA algorithm is strongly correlated with the degree to which the explaining-away matrix matches the ISTA algorithm.  Figure~\ref{incoming_part_figure} shows the decoders associated with the strongest recurrent connections to three representative part-units.  As expected, the decoders of these afferent units tend to be strongly aligned or anti-aligned with their target's decoder, and include both part-units and categorical-units.

\subsection{Categorical-units}

\begin{figure}[tb] 
  \begin{center}
    \begin{tabular}{p{0.02in}p{5.25in}}
      & \hspace{-0.1in} \begin{tabular}{p{0.175in}p{4in}} Src & Destination units \end{tabular} \\
      \parbox[b]{0in}{(a) \vspace{1.1cm}} & \includegraphics[width=5.2in, keepaspectratio=true]{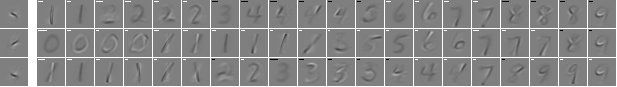} \\
      \parbox[b]{0in}{(b) \vspace{1.1cm}} & \includegraphics[width=5.2in, keepaspectratio=true]{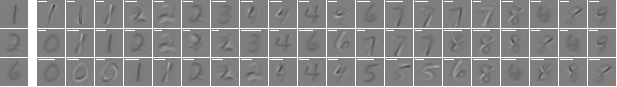} \vspace{0.1in} \\
      & \hspace{-0.1in} \begin{tabular}{p{0.175in}p{4in}} Dest & Source units \end{tabular} \\
      \parbox[b]{0in}{(c) \vspace{1.1cm}} &  \includegraphics[width=5.2in, keepaspectratio=true]{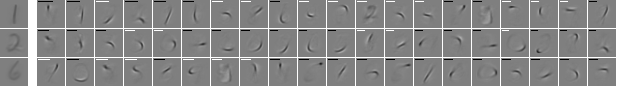} \\
    \end{tabular}
  \end{center}
  \caption{Categorical-units execute a sophisticated form of pooling 
    over part-units, and have winner-take-all dynamics amongst themselves.  The decoders of the categorical-units receiving the twenty strongest connections $\left|S_{i,j} - \delta_{i,j}\right|$ from representative part-units (a) and categorical-units (b), and the decoders of the part-units sending the twenty strongest projections to representative categorical-units (c).  The connections are sorted first by the class of their destination, and then by the magnitude of the connection.  The left-most column depicts the decoder of the source (a,b) or destination (c) unit.  The bars above the decoders in the remaining columns indicate the strength of the connections.  Black bars are used for positive connections, and white bars for negative connections.  
\label{categorical_connection_figure}}
\end{figure}

\begin{figure}[tb] 
  \begin{center}
    \begin{tabular}{p{0.04in}p{2.4in}p{0.04in}p{2.4in}}
      \parbox[b]{0in}{(a) \vspace{3.8cm}} & \includegraphics[width=2.5in, keepaspectratio=true]{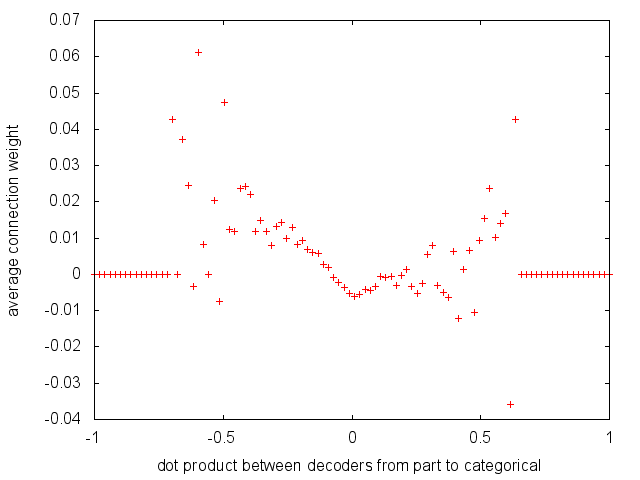} &
      \parbox[b]{0in}{(b) \vspace{3.8cm}} & \includegraphics[width=2.5in, keepaspectratio=true]{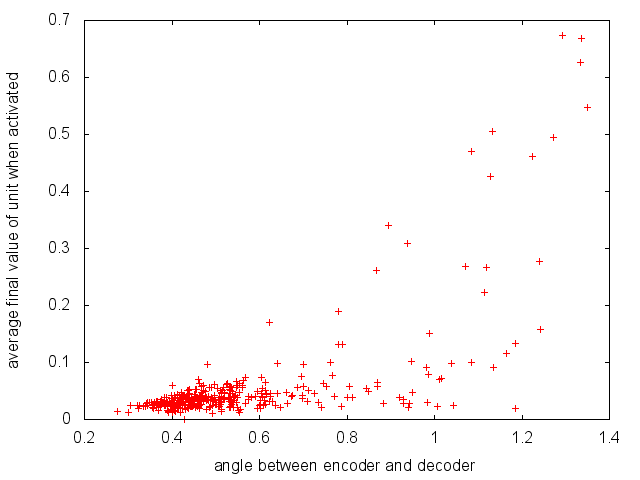} \\
      \parbox[b]{0in}{(c) \vspace{3.8cm}} & \includegraphics[width=2.5in, keepaspectratio=true]{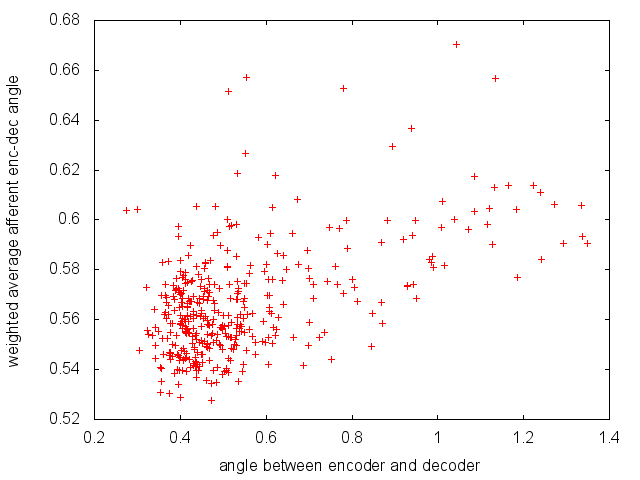} &
      \parbox[b]{0in}{(d) \vspace{3.8cm}} & \includegraphics[width=2.5in, keepaspectratio=true]{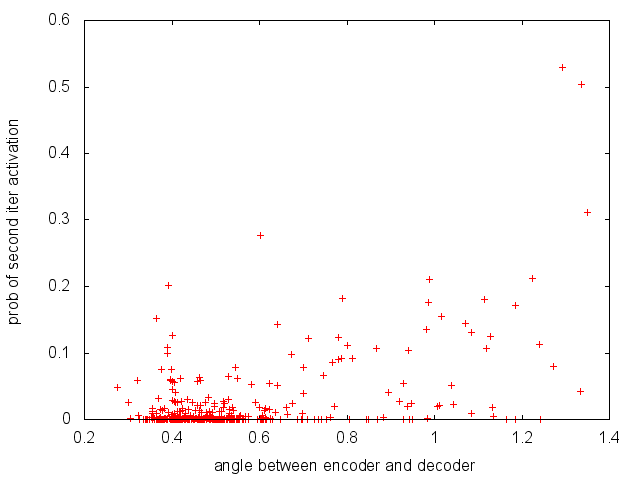} \\
      \parbox[b]{0in}{(e) \vspace{3.8cm}} & \includegraphics[width=2.5in, keepaspectratio=true]{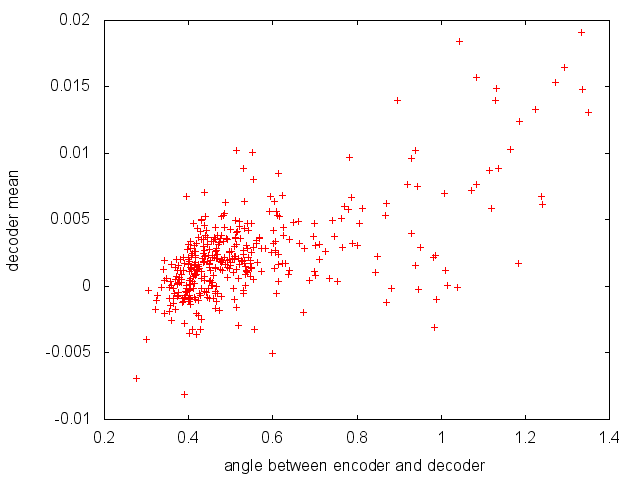}
    \end{tabular}
  \end{center}
  \caption{Statistics of connections indicate the presence of a rough hierarchy, with categorical-units on the top integrating 
    over part-units on the bottom. Average explaining-away connection weight $S_{ij}$, binned by alignment between decoders, for connections from part-units to categorical-units~(a).  If no units fall in a given bin, the average is set to zero. Average final value of a unit $\nobfhid_i^{t = T}$, given that $\nobfhid_i^{t=T} > 0$, versus the angle between the encoder row $E_i$ and decoder column $D_i$~(b).  Average angle between encoder row $E_j$ and decoder column $D_j$ of afferents to unit $i$, weighted by the strength of the connection to unit $i$, versus the angle between encoder row $E_i$ and decoder column $D_i$~(c). Probability that $\nobfhid_i^1 = 0$ and $\nobfhid_i^2 > 0$, versus the angle between the encoder row $E_i$ and decoder column $D_i$~(d).  Average value of the decoder column $\overline{D_i}$ versus the angle between the encoder row $E_i$ and the decoder column $D_i$~(e). \label{categorical_connection_statistics_figure}}  
\end{figure}

\begin{figure}[tb] 
  \begin{center}
    \begin{tabular}{p{0.12in}p{0.35in}p{4.05in}p{1in}} & Enc & Optimal inferred decoders & Dec \end{tabular} \\
    \begin{tabular}{p{0.03in}p{5.0in}}
      \parbox[b]{0in}{(a) \vspace{2.1cm}} & \includegraphics[width=5.1in, keepaspectratio=true]{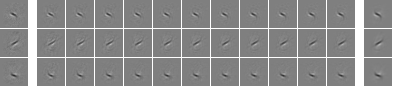} \\
      \parbox[b]{0in}{(b) \vspace{2.1cm}} & \includegraphics[width=5.1in, keepaspectratio=true]{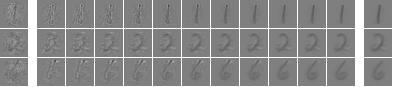} 
    \end{tabular}
  \end{center}
  \caption{Part-units (a) respond to the input quickly, while the activity of categorical-units (b) refines slowly.  Columns of the optimal decoding matrices $\D^t$ minimizing the input reconstruction error $\left|\left| \inp - \D^t \cdot \hid^t \right|\right|_2^2$ from the hidden representation $\hid^t$ for $t = 1, \ldots, T$.  The first and last columns show the corresponding encoder and decoder for the chosen representative units.  Intermediate columns represent successive iterations $t$.  \label{dictionary_evolution_figure}} 
\end{figure}

In contrast, the recurrent connections to categorical-units with an encoder-decoder angle greater than $0.7$ radians are not strongly correlated with the values predicted by ISTA.  
Rather than analyzing connections to the categorical-units only based upon their destination, it is more informative to consider them organized by their source.
Part-units are compatible with categorical-units of certain classes,\footnote{Categorical-units have strong, sparse classification matrix projections, as shown in figures~\ref{dictionary_figure}(c) and~\ref{two_classes_figure}(e,f), and can be identified with the output class to which they have the strongest projection.} and not with others, as shown by figure~\ref{categorical_connection_figure}(a).  
Part-units generally have positive connections to categorical-units with parallel prototypes, independent of offset, and negative connections to categorical-units with orthogonal prototypes, as shown in figure~\ref{categorical_connection_statistics_figure}(a).  This corresponds to a sophisticated form of pooling~\citep{jarrett2009},  with a single categorical-unit drawing excitation from a large collection of parallel but not necessarily perfectly aligned part-units, as in figure~\ref{categorical_connection_figure}(c).  It is also suggestive of the standard Hubel and Wiesel model of complex cells in primary visual cortex \citep{hubel1962}.  ISTA would instead predict a connection proportional to the inner product, which is zero for orthogonal prototypes and negative for anti-aligned prototypes. 

Part-units use sparse coding dynamics, and so are not disproportionately suppressed by categorical-units that represent any particular class.  However, each part-unit is itself compatible with (i.e., has positive connections to) categorical-units of only a subset of the classes.  As a result, the categorical-units and thus the class chosen are determined by the part-unit activations.  In particular, only a subset of the possible deformations implemented by part-unit decoders are freely available for each prototype, since part-units with a strong negative connection to a categorical-unit will tend to silence it, and so cannot be used to transform the prototype of that categorical-unit.

Categorical-units implement winner-take-all-like dynamics amongst themselves, as shown in figure~\ref{categorical_connection_figure}(b), with negative connections to most other categorical-units.  
Positive total self-connections $S_{i,i}$ facilitate the integration of inputs 
over time.

When activated, the categorical-units make a much larger contribution to the reconstruction than any single part-unit, as can be seen in figure~\ref{categorical_connection_statistics_figure}(b).  Since, the projections from categorical-units to part-units are consistent with ISTA, the magnitude of the categorical-unit contribution to the reconstruction need not be tightly regulated.  The part-units adjust accordingly to accommodate whatever residual is left by the categorical-units.

The units form a rough hierarchy, with part-units on the bottom and categorical-units on the top.  
Categorical-units receive strong recurrent connections, as shown in figure~\ref{two_classes_figure}(c,d) implying that their activity is more determined by other hidden units and less by the input (since the magnitude of the input connections is bounded), and thus they are higher in the hierarchy.  
As shown in figure~\ref{categorical_connection_statistics_figure}(c), part-units receive most of their input from other part-units; categorical-units receive a larger fraction of their input from other categorical-units.  
Whereas part-units have well-structured encoders and are generally activated directly by the input on the first iteration, categorical-units are more likely to first achieve a non-zero activation on the second iteration, as shown in figure~\ref{categorical_connection_statistics_figure}(d), suggesting that they require stimulation from part-units.  The immediate response of part-units in contrast to the gradual refinement of categorical-units is apparent in figure~\ref{dictionary_evolution_figure}, which shows the optimal decoding matrix for selected units, inferred from their observed activity at each iteration.  

\section{Performance}

\begin{table}[tb]
  \begin{center}
    \begin{tabular}{p{2.7in}p{0.25in}p{2.1in}} \toprule
      LISTA auto-encoder, $10 \times \left( 289 \text{-} 100^5 \right)$ & $3.76$ & ($5.98$ with 289 hidden units) \\ 
      \citep{sprechmann2012a} \\ \addlinespace[0.1cm] 
      Learned coordinate descent, $784 \text{-} 784^{50} \text{-} 10$ & $2.29$ \\ 
      \citep{gregor2010} \\ \addlinespace[0.1cm] 
      Differentiable sparse coding, $180 \text{-} 256^* \text{-} 10$ & $1.30$ \\ 
      \citep{bradley2008} \\ \addlinespace[0.1cm] %
      Deep sparse rectifier neural network &  $1.20$ &  ($1.16$ with tanh nonlinearity) \\ 
      $784$-$1000$-$1000$-$1000$-$10$ \\ \citep{glorot2011} \\ \addlinespace[0.1cm] 
      Deep belief network, $784 \text{-} 500 \text{-} 500 \text{-} 2000 \text{-} 10$ & $1.18$ & ($0.92$ with dropout) \\ 
      \citep{hinton2012} \\ \addlinespace[0.1cm] %
      {\bf Discriminative recurrent sparse auto-encoder} & $\mathbf{1.08}$ & ($1.21$ with 200 hidden units) \\ 
      $\mathbf{784}$-$\mathbf{400^{11}}$-$\mathbf{10}$ \\  \addlinespace[0.1cm] %
      Supervised dictionary learning, & $1.05$ & ($3.56$ without contrastive loss) \\ 
      $45 \times \left(784 \text{-} 24^* \right)$ to $45 \times \left( 784 \text{-} 96^* \right)$ \\
      \citep{mairal2009} \\ \addlinespace[0.1cm] \bottomrule 
    \end{tabular}
    \caption{MNIST classification error rate (\%) for pixel-permutation-agnostic encoders without boosting-like augmentations. The first column indicates the size of each layer in the specified encoder, separated by hyphens.  Exponents specify the number of recurrent iterations; asterisks denote repetition to convergence. $10 \times \left(\cdots\right)$ indicates that a separate encoder is trained for each input class; $45 \times \left(\cdots\right)$ indicates that a separate encoder is trained for each pairwise binary classification problem.  Further performance improvements have been reported with regularization techniques such as dropout, architectures that enforce translation-invariance, and datasets augmented by deformations, as discussed in the main text. 
\label{performance_table}} 
  \end{center}
\end{table}

The comparison of MNIST classification performance in table~\ref{performance_table} demonstrates the power of the hierarchical 
representation learned by DrSAEs. 
Rather than learn to minimize the sum of equations~\ref{reconstruction-loss} and~\ref{discriminative-loss}, \citet{gregor2010} train the LISTA encoder to approximate the code generated by a traditional sparse coder. 
While they do not report classification performance using LISTA, Gregor and LeCun do evaluate MNIST classification error using the related learned coordinate descent algorithm.
\citet{sprechmann2012a, sprechmann2012b} extend this approach by training a LISTA auto-encoder to reconstruct the input directly, using loss functions similar to equation~\ref{reconstruction-loss}. 
Although they identify the possibility of using regularization dependent upon supervised information, Sprechmann and colleagues do not consider a parameterized classifier operating on a common hidden representation. 
Instead, they train a separate encoder for each class, and classify each input based upon the encoder with the lowest sparse coding error.  
DrSAEs significantly outperform these other techniques based upon a LISTA encoder.

DrSAEs also perform well compared to other techniques using encoders related to LISTA.  Deep sparse rectifier neural networks~\citep{glorot2011} combine discriminative training with an encoder similar to LISTA, but do not tie the parameters between the layers and only allow the input to project to the first layer.  
Differentiable sparse coding~\citep{bradley2008} and supervised dictionary learning~\citep{mairal2009} also train discriminatively, but effectively use an infinite-depth ISTA-like encoder, and are thus much less computationally efficient than DrSAEs.  Supervised dictionary learning achieves performance statistically indistinguishable from DrSAEs using a contrastive loss function. 
A similar technique achieves MNIST classification error as low as $0.54\%$ when the dataset is augmented with shifted copies of the inputs~\citep{mairal2012}.

Additional regularizations and boosting-like techniques can further improve performance of networks with LISTA-like encoders. 
Recent examples include dropout, which trains and then averages over a large set of random subnetworks formed by removing a constant fraction of the hidden units from the original network \citep{goodfellow2013, hinton2012}. 
Deep belief networks and deep Boltzmann machines fine-tuned with dropout are the current state-of-the-art for pixel-permutation-agnostic handwritten digit recognition~\citep{hinton2012}, and can achieve MNIST classification error as low as $0.79\%$ with a carefully tuned network structure and multi-step training procedure.  
Deep convex networks, which iteratively refine the classification by successively training a stack of classifiers, with the output of the $i-1$st classifier provided as input to the $i$th classifier, can achieve an MNIST error of $0.83\%$ \citep{deng2011}.
Regularizing by explicit modeling of the data manifold, and then minimizing the square of the Jacobian of the output along the tangent bundle around the training datapoints, can reduce MNIST error to $0.81\%$ \citep{rifai2011b}.
Further performance improvements are possible if translation invariance is built directly into the network via a convolutional architecture, and deformations of the inputs are included in the training set \citep{lecun1998}, yielding error as low as $0.23 \%$ \citep{ciresan2012}.  These regularizations and augmentations are potentially compatible with DrSAE, but we defer their exploration to future work.  

Recurrence is essential to the performance of DrSAEs.  If the number of recurrent iterations is decreased from eleven to two, MNIST classification error in a network with 400 hidden units increases from $1.08\%$ to $1.32\%$.  With only 200 hidden units, MNIST classification error increases from $1.21\%$ to $1.49\%$, although the hidden units still differentiate into part-units and categorical-units, as shown in figure~\ref{two_classes_figure}(d,f).

\section{Discussion}

\begin{figure}[tb]
  \begin{center}
    \begin{tabular}{p{0.02in}p{5.25in}}
      &     \begin{tabular}{p{4.42in}p{0.15in}p{0.25in}} \hspace{-0.1in} Progressive reconstruction & Fin & Inp \end{tabular} \\
      \parbox[b]{0in}{(a) \vspace{3.1cm}} & \includegraphics[width=5.2in, keepaspectratio=true]{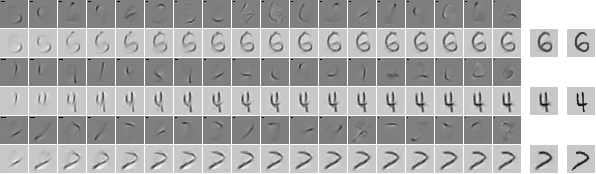} \\
      \parbox[b]{0in}{(b) \vspace{3.1cm}} & \includegraphics[width=5.2in, keepaspectratio=true]{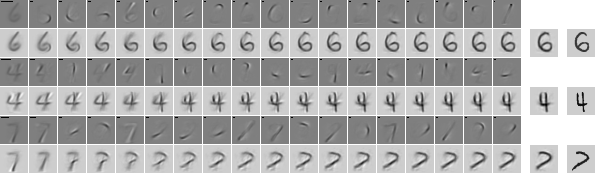} \\
      \parbox[b]{0in}{(c) \vspace{3.1cm}} & \includegraphics[width=5.2in, keepaspectratio=true]{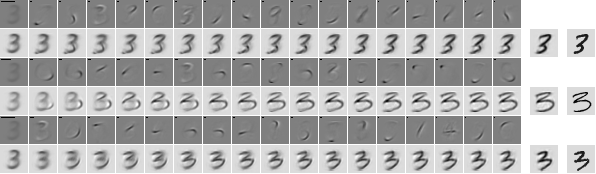} 
    \end{tabular}
  \end{center}
  \caption{Discriminative recurrent sparse auto-encoders decompose the input into a prototype and deformations along the data manifold. 
The progressive reconstruction of selected inputs by the hidden representation before (a) or after (b,c) discriminative fine-tuning.  
The columns from left to right depict either the components of the reconstruction (top row of each pair), or the partial reconstruction induced by the first $n$ parts (bottom row of each pair).  Parts are added to the reconstruction in order of decreasing contribution magnitude; smoother transformations are possible with an optimized sequence.  The last two columns show the final reconstruction with all parts (Fin), and the original input (Inp). Bars above the decoding matrix columns indicate the scale factor/hidden unit activity associated with the column. \label{gradual_reconstruction_figure}}
\end{figure}

\begin{figure}[tb]
  \begin{center}
  \begin{tabular}{p{4in}}
    \hspace{-0.1in} \begin{tabular}{p{0.22in}p{1.6in}p{0.22in}p{1.4in}} Avg & Most class-specific units & Avg & Most class-specific units \end{tabular} \\
    \includegraphics[width=4in, keepaspectratio=true]{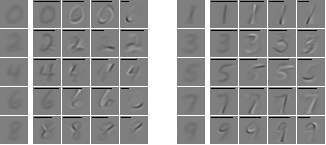} 
  \end{tabular}
  \end{center}
  \caption{The prototypes learned by categorical-units resemble representative instances of the appropriate class, and are sharper than the average over all members of the class in the dataset.  The left-most column in each group depicts the average over all elements of each of the ten MNIST digit classes.  The other columns show the decoders of the associated units with the largest-magnitude columns in the classification matrix $\C$.  Bars above the decoders indicate the angle between the encoder and the decoder for the displayed unit.  The most prototypical unit always makes the strongest contribution to the classification, and has a large (but not necessarily the largest) angle between its encoder and decoder.  Some units that make large contributions to the classification represent global transformations, such as rotations, of a prototype~\citep{simard1998}. \label{most_categorical_filters_figure}}
\end{figure}

It is widely believed that natural stimuli, such as images and sounds, fall near a low-dimensional manifold within a higher-dimensional space (the \emph{manifold hypothesis}) \citep{lee2003, olshausen2004, bengio2012b}.  
The low-dimensional data manifold provides an intuitively compelling and empirically effective basis for classification \citep{simard1993, simard1998, rifai2011b}.  The continuous deformations that define the data manifold usually preserve identity, whereas even relatively small invalid transformations may change the class of a stimulus.  For instance, the various handwritten renditions of the digit $3$ in in the last column of figure~\ref{gradual_reconstruction_figure}(c) barely overlap, and so the Euclidean distance between them in pixel space is greater than that to the nearest $8$ formed by closing both loops.  Nevertheless, smooth deformations of one $3$ into another correspond to relatively short trajectories along the data manifold,\footnote{In particular, figure~\ref{gradual_reconstruction_figure}(c) shows how each input can be produced by identity-preserving deformations from a common prototype, using the tangent space decomposition produced by our network.} whereas the transformation of a $3$ into an $8$ requires a much longer path within the data manifold.  
A prohibitive amount of data is required to fully characterize the data manifold \citep{narayanan2010}, so it is often approximated by the set of linear submanifolds tangent to the data manifold at the observed datapoints, known as the \emph{tangent spaces} \citep{simard1998, rifai2011b, ekanadham2011}.  
DrSAEs naturally and efficiently form a tangent space-like representation, consisting of a point on the data manifold indicated by the categorical-units, and a shift within the tangent space specified by the part-units.  

Before discriminative fine-tuning, DrSAEs perform a traditional part-based decomposition, familiar from sparse coding, as shown in figure~\ref{gradual_reconstruction_figure}(a).  The decoding matrix columns are class-independent, local pen strokes, and many units make a comparable, small contribution to the reconstruction.  After discriminative fine-tuning, the hidden units differentiate into sparse coding local part-units, and global prototype categorical-units that integrate 
over them.  As shown in figure~\ref{gradual_reconstruction_figure}(b,c), the input is decomposed into a prototype, corresponding to a point on the data manifold; and 
a set of deformations from this prototype along the data manifold, corresponding to shifts within the tangent space.  The same prototype can be used for very different inputs, as demonstrated in figure~\ref{gradual_reconstruction_figure}(c), since the space of deformations is rich enough to encompass diverse transformations without moving off the data manifold.  
Even when the prototype is very different from the input, all steps along the reconstruction trajectories in figure~\ref{gradual_reconstruction_figure}(b,c) are recognizable as members of the same class.

The prototypes learned by the categorical-units for each class are not simply the average over the elements of the class, as depicted in figure~\ref{most_categorical_filters_figure}.  Each class includes many possible input variations, so its average is blurry.  The prototypes, in contrast, are sharp, and look like representative elements of the appropriate class.  Many categorical-units are available for each class, as shown in figure~\ref{categorical_connection_figure}.  Not all categorical-units correspond to full prototypes; some capture global transformations of a prototype, such as rotations~\citep{simard1998}.

Consistent with prototypes for the non-negative MNIST inputs, the decoding matrix columns of the categorical-units are generally positive, as shown in figure~\ref{categorical_connection_statistics_figure}(e).  In contrast, the decoders of the part-units are approximately mean-zero and so cannot serve as prototypes themselves.  Rather, they shift and transform prototypes, moving activation from one region in the image to another, as demonstrated in figure~\ref{gradual_reconstruction_figure}(b,c).


Discrepancies between the prototype and the input due to transformations along the data manifold are explained by class-consistent part-units, and only serve to further activate the categorical-units of that class, as in figure~\ref{categorical_connection_figure}(a,c).
Discrepancies between the prototype and the input due to deformations orthogonal to the data manifold are explained by class-incompatible part-units, and serve to suppress the categorical-units of that class, both directly and via activation of incompatible categorical-units.  

If the wrong prototype is turned on, the residual input will generally contain substantial unexplained components.  Part-units obey ISTA-like dynamics and thus function as a sparse coder on the residual input, so part-units that match the unexplained components of the input will be activated.  These part-units will have positive connections to categorical-units with compatible prototypes, and so will tend to activate categorical-units associated with the true class (so long as the unexplained components of the input are diagnostic).  The spuriously activated categorical-unit will not be able to sustain its activity, since few compatible part-units will be required to capture the residual input.

The classification approach used by DrSAEs is different from one based upon a traditional sparse coding decomposition: it projects into the space of deviations from a prototype, which is not the same as the space of prototype-free parts, as is clear from figure~\ref{gradual_reconstruction_figure}(a,b).  For instance, a $5$ can easily be constructed using the parts of a $6$, making it difficult to distinguish the two.  Indeed, the first seven progressive reconstruction steps of the $6$ in figure~\ref{gradual_reconstruction_figure}(a) could just as easily be used to produce a $5$.  However, starting from a $6$ prototype, the parts required to break the bottom loop are outside the data manifold of the $6$ class, and so will tend to change the active prototype.  

DrSAEs naturally learn a hierarchical representation within a recurrent network, thereby implementing a deep network with parameter sharing between the layers.

\appendix


\begin{thebibliography}{100}
\providecommand{\natexlab}[1]{#1}
\expandafter\ifx\csname urlstyle\endcsname\relax
  \providecommand{\doi}[1]{doi:\discretionary{}{}{}#1}\else
  \providecommand{\doi}{doi:\discretionary{}{}{}\begingroup
  \urlstyle{rm}\Url}\fi

\addcontentsline{toc}{chapter}{Bibliography}


\bibitem[{Bengio(2009)}]{bengio2009}
Bengio, Y. (2009). 
\newblock Learning deep architectures for AI. 
\newblock \emph{Foundations and Trends in Machine Learning}, \emph{2}(1), 1--127.

\bibitem[{Bengio, Boulanger-Lewandowski, \& Pascanu(2012)}]{bengio2012a}
Bengio, Y., Boulanger-Lewandowski, N., \& Pascanu, R. (2012).
\newblock Advances in optimizing recurrent networks.
\newblock arXiv:1212.0901v2 [cs.LG] 

\bibitem[{Bengio, Courville, \& Vincent(2012)}]{bengio2012b}
Bengio, Y., Courville, A., \& Vincent, P. (2012).
\newblock Representation learning: A review and new perspectives.
\newblock arXiv:1206.5538 [cs.LG]

\bibitem[{Bengio \& Gingras(1996)}]{bengio1996}
Bengio, Y., \& Gingras, F. (1996).
\newblock Recurrent neural networks for missing or asynchronous data.
\newblock In D.~Touretzky, M.~Mozer, \& M.~Hasselmo (Eds.) \emph{Advances in Neural Information Processing Systems (NIPS 8)} (pp. 395--401).



\bibitem[{Bradley \& Bagnell(2008)}]{bradley2008}
Bradley, D. M., \& Bagnell, J. A. (2008).
\newblock Differentiable sparse coding.
\newblock In D. Koller, D. Schuurmans, Y. Bengio, \& L. Bottou (Eds.) \emph{Advances in Neural Information Processing Systems (NIPS 21)} (pp. 113--120).

\bibitem[{Boureau, et~al.(2010)Boureau, Bach, LeCun, \& Ponce}]{boureau2010}
Boureau, Y., Bach, F., LeCun, L., \& Ponce, J. (2010).
\newblock Learning mid-level features for recognition
\newblock In \emph{Proceedings of the 23rd IEEE Conference on Computer Vision and Pattern Recognition (CVPR 2010)}.

\bibitem[{Chambolle, et~al.(1998)Chambolle, De Vore, Lee, \& Lucier}]{chambolle1998}
Chambolle, A., De Vore, R. A., Lee, N. Y., \& Lucier, B. J. (1998). 
\newblock Nonlinear wavelet image processing: Variational problems, compression, and noise removal through wavelet shrinkage. 
\newblock \emph{IEEE Transactions on Image Processing}, \emph{7}(3), 319--335.

\bibitem[{Ciresan, Meier, \& Schmidhuber(2012)}]{ciresan2012}
Ciresan, D., Meier, U., \& Schmidhuber, J. (2012).
\newblock Multi-column deep neural networks for image classification.
\newblock In \emph{Proceedings of the 25th IEEE Conference on Computer Vision and Pattern Recognition (CVPR 2012)} (pp. 3642--3649).

\bibitem[{Coates \& Ng(2011)}]{coates2011}
Coates, A., \& Ng, A. Y. (2011).
\newblock The importance of encoding versus training with sparse coding and vector quantization
\newblock L. Getoor \& T. Scheffer (Eds.) \emph{Proceedings of the 28th International Conference on Machine Learning (ICML 2011)} (pp. 921--928).

\bibitem[{Dahl, et~al.(2012)Dahl, Yu, Deng, \& Acero}]{dahl2012}
Dahl, G. E., Yu, D., Deng, L., \& Acero, A. (2012). 
\newblock Context-dependent pre-trained deep neural networks for large-vocabulary speech recognition. 
\newblock \emph{IEEE Transactions on Audio, Speech, and Language Processing}, \emph{20}(1), 30--42.

\bibitem[{Daubechies, Defrise, \& De Mol(2004)}]{daubechies2004}
Daubechies, I., Defrise, M., \& De Mol, C. (2004). 
\newblock An iterative thresholding algorithm for linear inverse problems with a sparsity constraint. 
\newblock \emph{Communications on Pure and Applied Mathematics}, \emph{57}(11), 1413--1457.



\bibitem[{Ekanadham, Tranchina, \& Simoncelli(2011)}]{ekanadham2011}
Ekanadham, C., Tranchina, D., \& Simoncelli, E. P. (2011). 
\newblock Recovery of sparse translation-invariant signals with continuous basis pursuit.
\newblock \emph{IEEE Transactions on Signal Processing}, \emph{59}(10), 4735--4744.


\bibitem[{Glorot, Bordes, \& Bengio(2011)}]{glorot2011}
Glorot, X., Bordes, A., \& Bengio, Y. (2011).
\newblock Deep sparse rectifier neural networks. 
\newblock In G. Gordon, D. Dunson, \& M. Dudik (Eds.) \emph{JMLR W\&CP: Proceedings of the Fourteenth International Conference on Artificial Intelligence and Statistics (AISTATS 2011)} (pp. 315--323).

\bibitem[{Goodfellow, et~al.(2013)Goodfellow, Warde-Farley, Mirza, Courville, \& Bengio}]{goodfellow2013}
Goodfellow, I. J., Warde-Farley, D., Mirza, M., Courville, A., \& Bengio, Y. (2013).
\newblock Maxout networks
\newblock arXiv:1302.4389v3 [stat.ML]

\bibitem[{Gregor \& LeCun(2010)}]{gregor2010}
Gregor, K., \& LeCun, Y. (2010).
\newblock Learning fast approximations of sparse coding.
\newblock In J. F{\"u}rnkranz \& T. Joachims (Eds.) \emph{Proceedings of the 27th International Conference on Machine Learning (ICML 2010)} (pp. 399--406).


\bibitem[{Hinton, Osindero, \& Teh(2006)}]{hinton2006}
Hinton, G. E., Osindero, S., \& Teh, Y. W. (2006). 
\newblock A fast learning algorithm for deep belief nets. 
\newblock \emph{Neural Computation}, \emph{18}(7), 1527--1554.

\bibitem[{Hinton(2010)}]{hinton2010}
Hinton, G. (2010).
\newblock A practical guide to training restricted Boltzmann machines (UTML TR 2010-003, version 1).
\newblock Toronto, Canada: University of Toronto, Department of Computer Science.

\bibitem[{Hinton, et~al.(2012)Hinton, Srivastava, Krizhevsky, Sutskever, \& Salakhutdinov}]{hinton2012}
Hinton, G. E., Srivastava, N., Krizhevsky, A., Sutskever, I., \& Salakhutdinov, R. R. (2012). 
\newblock Improving neural networks by preventing co-adaptation of feature detectors
\newblock arXiv:1207.0580v1  [cs.NE]

\bibitem[{Hubel \& Wiesel(1962)}]{hubel1962}
Hubel, D. H., \& Wiesel, T. N. (1962). 
\newblock Receptive fields, binocular interaction and functional architecture in the cat's visual cortex. 
\newblock \emph{The Journal of Physiology}, \newblock{160}(1), 106--154.



\bibitem[{Jarrett, et~al.(2009)Jarrett, Kavukcuoglu, Ranzato, \& LeCun}]{jarrett2009}
Jarrett, K., Kavukcuoglu, K., Ranzato, M. A., \& LeCun, Y. (2009). 
\newblock \emph{What is the best multi-stage architecture for object recognition?} 
\newblock In \emph{Proceedings of the 12th International Conference on Computer Vision (ICCV 2009)} (pp. 2146--2153). 




\bibitem[{LeCun, et~al.(1998)LeCun, Bottou, Bengio, \& Haffner}]{lecun1998}
LeCun, Y., Bottou, L., Bengio, Y., \& Haffner, P. (1998). 
\newblock Gradient-based learning applied to document recognition. 
\newblock \emph{Proceedings of the IEEE}, \emph{86}(11), 2278--2324.

\bibitem[{Lee, Pedersen, \& Mumford(2003)}]{lee2003}
Lee, A. B., Pedersen, K. S., \& Mumford, D. (2003). 
\newblock The nonlinear statistics of high-contrast patches in natural images. 
\newblock \emph{International Journal of Computer Vision}, \emph{54}(1), 83--103.


\bibitem[{Lee, Ekanadham, \& Ng(2008)}]{lee2008}
Lee, H., Ekanadham, C., \& Ng, A. (2008). 
\newblock Sparse deep belief net model for visual area V2. 
\newblock In J. C. Platt, D. Koller, Y. Singer \& S. Roweis (Eds.) \emph{Advances in Neural Information Processing Systems (NIPS 20)}, (pp. 873--880).


\bibitem[{Mairal, et~al.(2009)Mairal, Bach, Ponce, Sapiro, \& Zisserman}]{mairal2009}
Mairal, J., Bach, F., Ponce, J., Sapiro, G., \& Zisserman, A. (2009).
\newblock Supervised dictionary learning.
\newblock In D. Koller, D. Schuurmans, Y. Bengio, \& L. Bottou (Eds.) \emph{Advances in Neural Information Processing Systems (NIPS 21)} (pp. 1033--1040).

\bibitem[{Mairal, Bach, \& Ponce(2012)}]{mairal2012}
Mairal, J., Bach, F., \& Ponce, J. (2012). 
\newblock Task-driven dictionary learning. 
\newblock \emph{IEEE Transactions on Pattern Analysis and Machine Intelligence}, \emph{34}(4), 791--804.


\bibitem[{Nair \& Hinton(2010)}]{nair2010}
Nair, V., \& Hinton, G. E. (2010). 
\newblock Rectified linear units improve restricted boltzmann machines. 
\newblock In J. F{\"u}rnkranz \& T. Joachims (Eds.) \emph{Proceedings of the 27th International Conference on Machine Learning (ICML 2010)} (pp. 807-814). 

\bibitem[{Narayanan \& Mitter(2010)}]{narayanan2010}
Narayanan, H. \& MItter, S. (2010).
\newblock Sample complexity of testing the manifold hypothesis.
\newblock In J. Lafferty, C. K. I. Williams, J. Shawe-Taylor, R.S. Zemel, \& A. Culotta (Eds.) \emph{Advances in Neural Information Processing Systems (NIPS 23)} (pp. 1786--1794).

\bibitem[{Olshausen \& Field(1996)}]{olshausen1996}
Olshausen, B. A., \& Field, D. J. (1996). 
\newblock Emergence of simple-cell receptive field properties by learning a sparse code for natural images. 
\newblock \emph{Nature}, \emph{381}(6583), 607--609.

\bibitem[{Olshausen \& Field(1997)}]{olshausen1997}
Olshausen, B. A., \& Field, D. J. (1997). 
\newblock Sparse coding with an overcomplete basis set: A strategy employed by VI?
\newblock \emph{Vision Research}, \emph{37}(23), 3311--3326.

\bibitem[{Olshausen \& Field(2004)}]{olshausen2004}
Olshausen, B. A., \& Field, D. J. (2004). 
\newblock Sparse coding of sensory inputs. 
\newblock \emph{Current opinion in neurobiology}, \emph{14}(4), 481--487.

\bibitem[{Ranzato, et~al.(2006)Ranzato, Poultney, Chopra, \& LeCun}]{ranzato2006}
Ranzato M., Poultney, C., Chopra, S., \& LeCun, Y. (2006). 
\newblock Efficient learning of sparse representations with an energy-based model.
\newblock In B. Sch\"{o}lkopf, J. Platt, \& T. Hoffman (Eds.) \emph{Advances in Neural Information Processing Systems (NIPS 19)}, (pp. 1137--1144).

\bibitem[{Ranzato \& Szummer(2008)}]{ranzato2008}
Ranzato, M., \& Szummer, M. (2008). 
\newblock Semi-supervised learning of compact document representations with deep networks
\newblock In A. McCallum \& S. Roweis (Eds.), \emph{Proceedings of the 25th Annual International Conference on Machine Learning (ICML 2008)} (pp. 792--799).



\bibitem[{Rifai, et~al.(2011)Rifai, Dauphin, Vincent, Bengio, \& Muller}]{rifai2011b}
Rifai, S., Dauphin, Y., Vincent, P., Bengio, Y., \& Muller, X. (2011). 
\newblock The manifold tangent classifier. 
\newblock In J. Shawe-Taylor, R. S. Zemel, P. Bartlett, F. C. N. Pereira, \& K. Q. Weinberger (Eds.) \emph{Advances in Neural Information Processing Systems (NIPS 24)} (pp. 2294--2302). 


\bibitem[{Rumelhart, et~al.(1986)Rumelhart, Hinton, \& Williams}]{rumelhart1986}
Rumelhart, D. E., Hinton, G. E., \& Williams, R. J. (1986).
\newblock Learning internal representations by error propagation.
\newblock In D. E. Rumelhart, J. L. McClelland, and the PDP Research Group (Eds.), \emph{Parallel Distributed Processing: Explorations in the Microstructure of Cognition: Vol. 1. Foundations} (pp. 318--362).  Cambridge, MA: MIT Press.  

\bibitem[{Salinas \& Abbott(1996)}]{salinas1996}
Salinas, E., \& Abbott, L. F. (1996). 
\newblock A model of multiplicative neural responses in parietal cortex. 
\newblock \emph{Proceedings of the National Academy of Sciences of the United States of America}, \emph{93}(21), 11956--11961.


\bibitem[{Seung(1998)}]{seung1998}
Seung, H. S. (1998).
\newblock Learning continuous attractors in recurrent networks.
In M. I. Jordan, M. J. Kearns, \& S. A. Solla (Eds.) \emph{Advances in Neural Information Processing Systems (NIPS 10)} (pp. 654--660). 

\bibitem[{Simard, LeCun, \& Denker(1993)}]{simard1993}
Simard, P., LeCun, Y., \& Denker, J. S. (1993). 
\newblock Efficient pattern recognition using a new transformation distance. 
In S. J. Hanson, J. D. Cowan, \& C. L. Giles (Eds.) \emph{Advances in Neural Information Processing Systems (NIPS 5)} (pp. 50--58). 

\bibitem[{Simard, et~al.(1998)Simard, LeCun, Denker, \& Victorri}]{simard1998}
Simard, P., LeCun, Y., Denker, J., \& Victorri, B. (1998). 
\newblock Transformation invariance in pattern recognition: Tangent distance and tangent propagation. 
\newblock In G. Orr,  \& K. Muller (Eds.), \emph{Neural networks: Tricks of the trade}. Berlin: Springer. 


\bibitem[{Sprechmann, Bronstein, \& Sapiro(2012a)}]{sprechmann2012a}
Sprechmann, P., Bronstein, A., \& Sapiro, G. (2012).
\newblock Learning efficient structured sparse models.
\newblock In J. Langford \& J. Pineau (Eds.) \emph{Proceedings of the 29th International Conference on Machine Learning (ICML 12)} (pp. 615--622).

\bibitem[{Sprechmann, Bronstein, \& Sapiro(2012b)}]{sprechmann2012b}
Sprechmann, P., Bronstein, A., \& Sapiro, G. (2012).
\newblock Learning efficient sparse and low rank models.
\newblock arXiv:1212.3631 [cs.LG]



\bibitem[{Deng \& Yu(2011)}]{deng2011}
Deng, L. \& Yu, D. (2011).
\newblock Deep convex net: A scalable architecture for speech pattern classification.
\newblock In \emph{Proceedings of the 12th Annual Conference of the International Speech Communication Association (INTERSPEECH 2011)} (pp. 2285-2288).



\bibitem[{Zeiler, Taylor, \& Fergus(2011)}]{zeiler2011}
Zeiler, M. D., Taylor, G. W., \& Fergus, R. (2011).
\newblock Adaptive deconvolutional networks for mid and high level feature learning.
\newblock In \emph{Proceedings of the 13th International Conference on Computer Vision (ICCV 2011)} (pp. 2018--2025).


\end{thebibliography}
\end{document}